\documentclass[nohyperref]{article}

\usepackage{microtype}
\usepackage{graphicx}
\usepackage{booktabs} 
\usepackage{array, multirow}
\usepackage{subcaption}

\usepackage{tabularx}
\usepackage{hyperref}
\usepackage{arydshln}

\usepackage{listings}

\usepackage{relsize}
\lstdefinestyle{textFileStyle}{
  basicstyle=\scriptsize\ttfamily,        
  breaklines=true,
backgroundcolor=\color{gray!10}, 
    frame=single, 
  numbers=none,                    
  breakindent=0pt,
  numbersep=5pt,                   
  numberstyle=\tiny\color{gray}, 
  rulecolor=\color{black},         
  showspaces=false,                
  showstringspaces=false,          
  showtabs=false,                  
  stringstyle=\color{purple},     
  tabsize=4,	                   
  title=\lstname,                   
  xleftmargin=.2cm,
  xrightmargin=.2cm,
  aboveskip=1.2em,
  belowskip=-1.5 \baselineskip,
  belowcaptionskip=0em
  }

\lstdefinestyle{python}{ 
  backgroundcolor=\color{white},   
  basicstyle=\scriptsize\ttfamily,        
  breakatwhitespace=false,         
  breaklines=true,                 
  captionpos=b,                    
  commentstyle=\color{green},    
  escapeinside={\%*}{*)},          
  extendedchars=false,              
  frame=lrtb,	                   
  keepspaces=true,                 
  keywordstyle=\color{blue},       
  language=Python,
  morekeywords={with,as},
  numbers=none,                    
  numbersep=5pt,                   
  numberstyle=\tiny\color{gray}, 
  rulecolor=\color{black},         
  showspaces=false,                
  showstringspaces=false,          
  showtabs=false,                  
  stringstyle=\color{purple},     
  tabsize=4,	                   
  title=\lstname,                   
  xleftmargin=.2cm,
  xrightmargin=.2cm,
  aboveskip=1.2em,
  belowskip=-1.5 \baselineskip,
  belowcaptionskip=0em,
}

\usepackage{tikz}
\usepackage{pgfplots}
\pgfplotsset{compat=1.16}
\usepackage{subcaption}
\usepackage{xcolor}
\usepackage{enumitem}
\usepackage[font=small]{caption}
\usepackage{stfloats}
\usepackage{colortbl}
\usepackage{makecell}

\definecolor{dred}{HTML}{F25F5C}
\definecolor{dblue}{HTML}{247BA0}
\definecolor{dgreen}{HTML}{70C1B3}
\definecolor{darkdgreen}{HTML}{56897A} 
\definecolor{darkdblue}{HTML}{1B5B77}  

\definecolor{g-blue}{RGB}{66,133,244}
\definecolor{g-red}{RGB}{234,67,53}
\definecolor{g-yellow}{RGB}{251,188,4}
\definecolor{g-green}{RGB}{52,168,82}

\usepackage{tikz}
\usepackage{pifont}
\usepackage{xcolor,colortbl}
\usepackage{makecell}
\usetikzlibrary{plotmarks}
\usetikzlibrary{shapes, calc}
\usetikzlibrary{fit}
\usetikzlibrary{patterns}
\usetikzlibrary{positioning}

\usepackage[accepted]{icml2024}

\usepackage{amsmath}
\usepackage{amssymb}
\usepackage{mathtools}
\usepackage{amsthm}
\usepackage{booktabs}
\usepackage{siunitx}
\usepackage[capitalize,noabbrev]{cleveref}

\theoremstyle{plain}

\theoremstyle{definition}

\theoremstyle{remark}

\usepackage[textsize=tiny]{todonotes}

\newcommand{\ours}{\ensuremath{\mathrm{BAGEL}}}

\icmltitlerunning{BAGEL: Bootstrapping Agents by Guiding Exploration with Language}

\begin{document}

\twocolumn[

\icmltitle{BAGEL: Bootstrapping Agents by Guiding Exploration with Language}
\icmlsetsymbol{equal}{*}

\begin{icmlauthorlist}
\icmlauthor{Shikhar Murty\textsuperscript{$\star$}}{yyy}
\icmlauthor{Christopher D. Manning}{yyy}
\icmlauthor{Peter Shaw}{comp}
\icmlauthor{Mandar Joshi}{comp}
\icmlauthor{Kenton Lee}{comp}

\end{icmlauthorlist}

\icmlaffiliation{yyy}{Department of Computer Science, Stanford University}
\icmlaffiliation{comp}{Google Deepmind}

\icmlcorrespondingauthor{Shikhar Murty}{smurty@cs.stanford.edu}

\icmlkeywords{Machine Learning, ICML}

\vskip 0.3in
]

\printAffiliationsAndNotice{\icmlEqualContribution} 

\begin{abstract}
Following natural language instructions by executing actions in digital environments (e.g. web-browsers and REST APIs) is a challenging task for language model (LM) agents.
Unfortunately, LM agents often fail to generalize to new environments without human demonstrations. This work presents $\ours{}$, a method for bootstrapping LM  agents \emph{without human supervision}. $\ours{}$ converts a seed set of randomly explored trajectories or synthetic instructions, into demonstrations, via round-trips between two noisy LM components: an LM \emph{labeler} which converts a trajectory into a synthetic instruction, and a zero-shot LM agent which maps the synthetic instruction into a refined trajectory. By performing these round-trips iteratively, $\ours{}$ quickly converts the initial distribution of trajectories towards those that are well-described by natural language. We use $\ours{}$ demonstrations to adapt a zero shot LM agent at test time via in-context learning over retrieved demonstrations, and find improvements of over 2-13\% absolute on ToolQA and MiniWob++, with up to 13$\times$ reduction in execution failures.
\end{abstract}

\section{Introduction}
\label{submission}

\begin{figure}[!ht]
    \centering
    \includegraphics[width=0.45\textwidth]{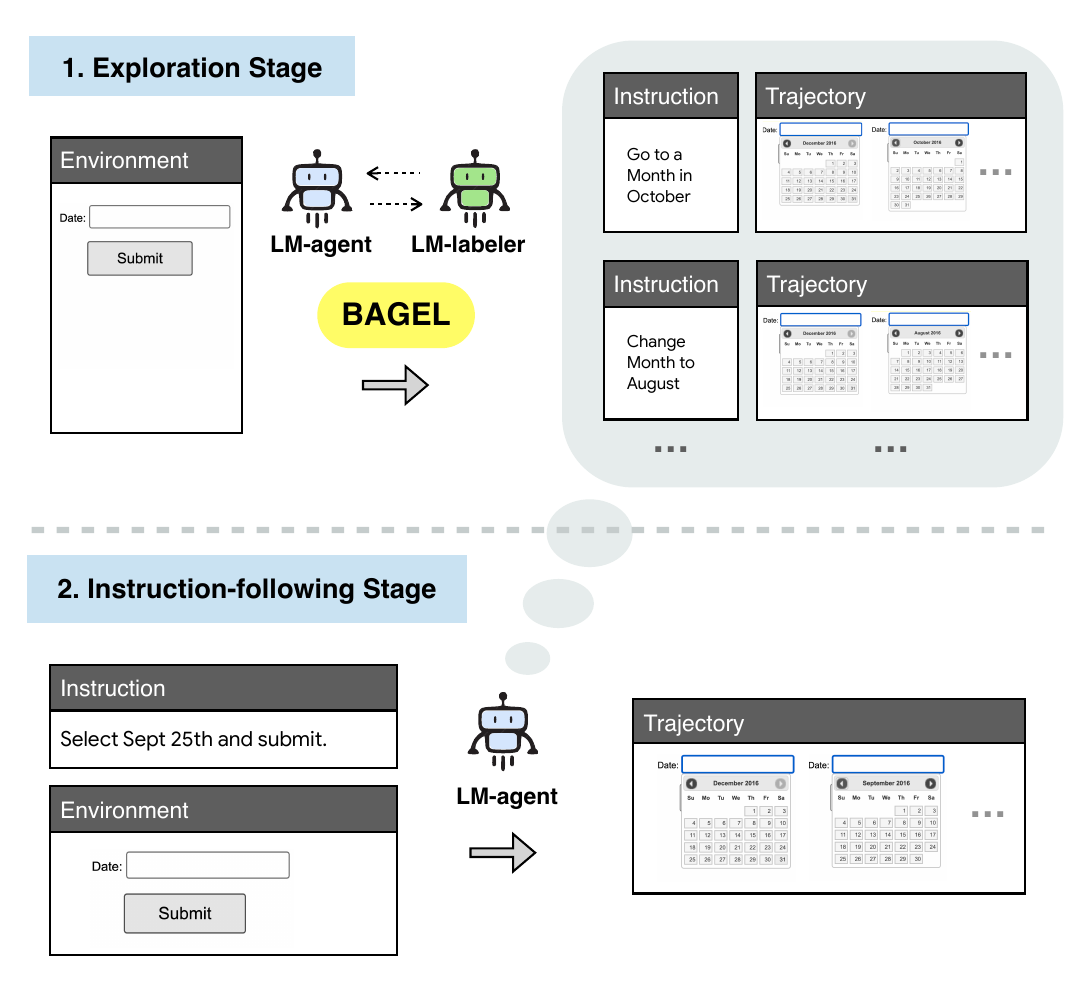}
    \caption{(Top) Given a seed set of explored trajectories, $\ours{}$ constructs synthetic demonstrations via an iterative round-trip procedure between two LM components: a zero-shot LM agent that generates trajectories and an LM labeler that generates instructions for these trajectories. (Bottom) Given an instruction at test time, we retrieve synthetic demonstrations with similar instructions, to use as in-context exemplars to adapt the base agent.}
    \label{fig:overview_fig}
\end{figure}
In recent years, large language models (LLMs) have shown strong performance on a broad range of language understanding tasks, making them powerful tools for controlling policies in digital environments such as web browsers \citep{yao2022react, kim2023language}. Such grounded language understanding tasks are fundamentally challenging for LMs in environments with ambiguous dynamics. For instance, even inputting a date into a text box could require either simply typing or a complex interaction using a drop-down date picker. An LM cannot know this a-priori without in-depth knowledge about the website.

One common way to provide such knowledge to LM agents is via expert demonstrations that provide information about mapping instructions to action sequences, recovering from errors, and reasoning traces~\cite{yao2022react, sun2023adaplanner, kim2023language, sodhi2023heap}. Of course, collecting human demonstrations for every new environment is laborious and requires knowing possible user instructions \emph{a priori}. Moreover, as agents scale to complex tasks with hundreds of actions, human supervision will become increasingly infeasible to obtain. Instead of relying on human demonstrations for training LM agents, could we instead use exploration and environment feedback to automatically collect a large number of \emph{synthetic} demonstrations?

Prior work has shown the effectiveness of collecting synthetic demonstrations by retroactively labeling trajectories from embodied agents~\citep{sumers2023distilling}. In this scenario, the environments dynamics are assumed to be well understood by the agent; the synthetic demonstrations only serve to connect agent behavior with human language. However, we observe the opposite challenge with digital agents in our setting---grounding instructions is relatively easy due to the highly textual environment, but \emph{zero-shot} digital agents typically are not exposed to any environment dynamics before they are directly used to follow instructions.

\begin{figure*}[!ht]
\begin{center}
\input{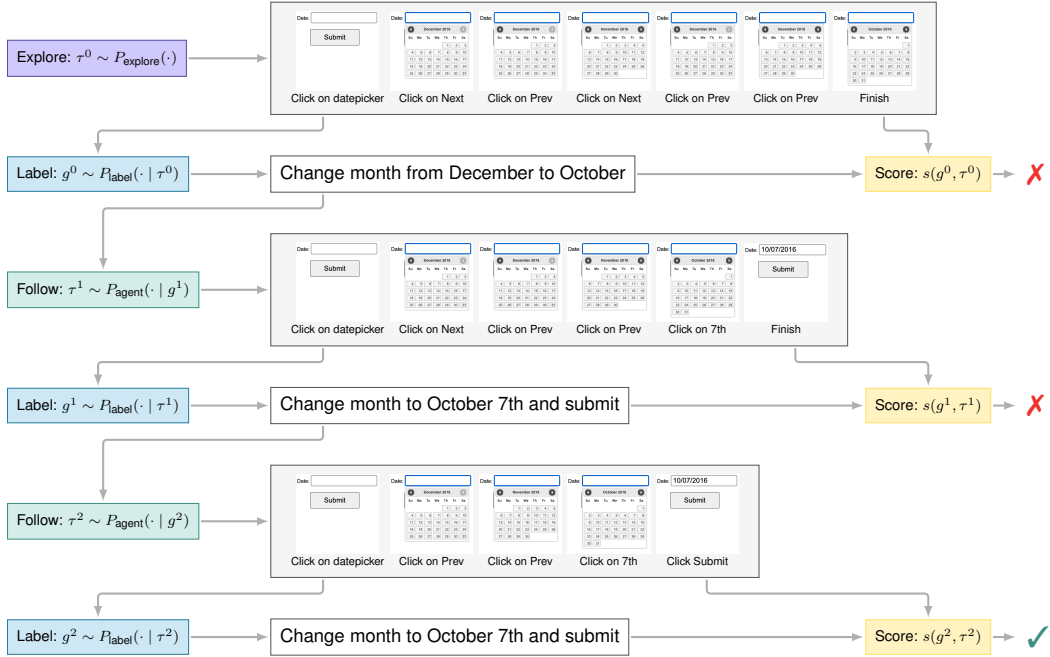}
\end{center}
\caption{\ours~generates synthetic demonstrations by exploring the environment. Shown here is an example from the MiniWob++ \textit{choose-date} task. First, we generate an initial trajectory by sampling actions without conditioning on any natural language instruction. Then, we alternate between generating an instruction given a trajectory, and generating a trajectory given an instruction. The process aims to converge towards a trajectory that accurately satisfies a natural language instruction, and aims to recover from errors in labeling or instruction following from earlier rounds (see example). Once an instruction and trajectory pair satisfies a filtering criteria, it is added to the set of synthetic demonstrations. Alternatively, \ours~can be initialized by first sampling an instruction, as described in \S\ref{sec:exploration}.}
\label{fig:overview_method}
\end{figure*}

Our method, termed $\ours{}$ (\textbf{B}ootstrapping \textbf{A}gents by \textbf{G}uiding \textbf{E}xploration with \textbf{L}anguage), uses an iterative procedure to relabel a seed set of trajectories obtained from unconditioned exploration (Figure~\ref{fig:overview_fig}). Intuitively, $\ours{}$ operates by progressively shifting the distribution of trajectories towards those that can be well-described via natural language, using two noisy LM components: an LM \textit{labeler} takes a trajectory and relabels it with a synthetic instruction, and a zero-shot LM policy maps the instruction back into a refined trajectory (Figure~\ref{fig:overview_method}). By performing these round trips iteratively, $\ours{}$ converts trajectories from random exploration into meaningful trajectories that are executable, without requiring a trained base agent or significant information about possible instructions. While both the re-labeling and instruction-following processes are imperfect, round-trips between these components work in harmony to reduce any noise. Once an instruction, trajectory pair reaches a threshold score under a \emph{demonstration filter} (another prompted LM), the generated synthetic demonstration is added into a buffer. $\ours{}$ demonstrations can be used for both in-context learning or finetuning, and serve as a drop-in replacement for expert demonstrations. Here, we follow the former strategy along with a simple retrieval augmented generation procedure---given a user instruction at test time,  we retrieve the most relevant demonstrations based on instruction embeddings, and feed that into the agent's prompt to serve as in-context exemplars.

While $\ours{}$ shares some similarities with Hindsight Experience Replay (HER, \citealp{andrychowicz2017hindsight}), a popular method for retroactive relabeling of unsuccessful trajectories, there are important technical differences: Instead of relabeling trajectories based on only the final observation, our relabeling function operates on the entire transition history from the trajectory and uses language models to \emph{iteratively} enforce a language prior over the distribution of trajectories ( Section~\ref{sec:discussion}). Moreover, while HER is used in offline Q-learning settings, we use $\ours{}$ primarily as a data generation method.

We experiment with $\ours{}$ on two domains, by using a prompted LM (similar to ReAct, \citealp{yao2022react}) as our base policy and find significant improvements with \emph{no human supervision}. In MiniWoB++ \citep{shi2017world, liu2018reinforcement}, an agent follows instructions on diverse web-interfaces ranging from booking flights to replying to emails, given an HTML state, by issuing a sequence of mouse and keyboard operations to interact with DOM objects. Using \ours{} for test-time adaptation, we find an improvement of over 13\% compared to the base LM policy. 
Next, we evaluate on ToolQA \citep{zhuang2023toolqa}, a collection of question answering tasks over 8 domains, where answering each question requires chaining together multiple tools such as SQL, text retrievers, graph tools, python interpreters and calculators. Here, we find an improvement of 2\% over the base LM policy. Further analysis reveals the various positive effects of conditioning on our synthetic demonstration beyond improved accuracy, including up to 13$\times$ reduction in execution failures due to better understanding of environment dynamics. By carefully using LM priors to shape random exploration, our method serves as a tool for automated discovery of use cases in complex environments. 

\section{Background}
Given a natural language instruction $g$, our agent interacts with the environment by taking a sequence of actions $\{a_1, a_2, \ldots, a_T\}$, where each $a_t$ is issued in response to an environment observation $o_t$. The entire interaction with the environment is captured as a \emph{trajectory} $\tau = \{o_1, a_1, o_2, \ldots, o_{T}, a_T, o_{T+1}\}$. 

We define an agent as a \emph{language conditioned policy} $\pi(a_t \mid \tau_{<t}, g)$ where $\tau_{<t} = \{o_1, a_1, o_2, \ldots, o_{t}\}$ refers to the trajectory until time-step $t$. Such policies are typically trained via imitation learning and optional RL finetuning, where a large set of expert curated instruction-trajectory pairs are required for imitation learning, and a suitably shaped reward signal is needed for RL finetuning \citep{branavan2009reinforcement, chaplot2018gated, misra2017mapping}.
For our setup, both observations and actions can be expressed as natural language strings. The agent policy $\pi$ can then be cast into an autoregressive LM that assigns probabilities to action strings given string descriptions of the previous actions and observations. Thus, recent work focuses on directly using LLMs as policies, by using prompts along with in-context human demonstrations \citep[among others]{yao2022react, shinn2023reflexion, sun2023adaplanner, kim2023language}.

\paragraph{Executing Action Strings.} 
Similar to prior work that uses LMs to generate action strings \citep{huang2022language, logeswaran2022few}, we assume access to an environment specific \emph{low-level controller} that maps action strings to a low-level command (e.g. a web-driver action or an API call), which can be directly executed to change the environment. 

\section{BAGEL}
$\ours{}$ generates synthetic demonstrations via exploration, as illustrated in Figure~\ref{fig:overview_method}. First, we describe the various model components in \S\ref{sec:model-components}, and then describe the overall procedure in \S\ref{sec:exploration}.

\subsection{Model Components}
\label{sec:model-components}
In order to generate synthetic demonstrations, we 
model different aspects of the joint distribution over instructions and trajectories. Every component is implemented by the same underlying LM, but with different prompts. Every component is also implicitly dependent on a given environment, although this is omitted in the notation for simplicity. All prompts used can be found in Appendix~\ref{sec:prompts}.

\paragraph{Exploration Policy.} The exploration policy, $\pi_\text{explore}(a_t \mid \tau_{<t})$, selects an action without conditioning on any instruction. The prompt used is similar to that of ReAct~\cite{yao2022react}. We can sample from the resulting distribution over trajectories, $p_\text{explore}(\tau)$, by sampling actions from $\pi_\text{explore}$ until the episode completes or a ``finish'' action is generated. We can increase the entropy of $\pi_\text{explore}$ with a configurable temperature parameter.

\paragraph{Trajectory Labeler.} The trajectory labeler, $p_\text{label}(g \mid \tau)$, is prompted to generate an instruction, $g$, that corresponds to a given trajectory, $\tau$.

\paragraph{Instruction Following Policy.} Unlike the exploration policy, the instruction following policy, $\pi_{\text{agent}}(a_t \mid \tau_{<t}, g)$, selects actions conditioned on an instruction, $g$. We sample from the resulting distribution over trajectories, $p_\text{agent}(\tau \mid g)$, by choosing actions according to $\pi_\text{agent}$ until the episode completes or a ``finish'' action is generated. This component is also implemented using a ReAct based prompt.

\paragraph{Demonstration Filter.} Given a synthetic demonstration $(g, \tau)$, the demonstration filter makes a binary judgement $s(g, \tau) \in \{0, 1\}$, based on how well $\tau$ corresponds to the instruction $g$.

\paragraph{Instruction Generator} Finally, as an alternative to the exploration policy (see \S\ref{sec:exploration}) we can instead use an instructor generator to initialize exploration. This model defines a distribution over instructions, $p_\text{instruct}(g)$, based on a prompt that elicits plausible instructions based on the initial observation from the environment, and the action space.

\subsection{Generating Demonstrations}
\label{sec:exploration}

\paragraph{Initial Exploration} We consider and compare two different variations of $\ours$: \textit{trajectory-first} and \textit{instruction-first} exploration. For trajectory-first exploration, we first sample a trajectory $\tau^0 \sim p_\text{explore}(\cdot)$ with the exploration policy. For instruction-first exploration, we first sample an instruction $g^0 \sim p_\text{instruct}(\cdot)$ with the instruction generator.

\paragraph{Iterative Refinement}
Trajectories sampled from $p_\text{explore}$ may not correspond to any reasonable instruction, and, similarly, there may be no feasible trajectory that satisfies instructions sampled from $p_\text{instruct}$. Our iterative re-labeling procedure aims to find an instruction and trajectory pair where the trajectory satisfies the instruction, without sacrificing the diversity of the initial exploration.
The process alternates between sampling instructions and trajectories:
\begin{align}
    g^{t} \sim p_\text{label}(\cdot \mid \tau^{t}).\\
    \tau^{t+1} \sim p_\text{agent}(\cdot \mid g^{t}).
\end{align}

We perform these iterative updates until we find a pair where $s(g^t, \tau^t) = 1$ or a maximum number of steps is reached. If we are successful, the demonstration $(g^t, \tau^t)$ is added to the set of synthetic demonstrations, $\mathcal{M}$. The overall procedure is repeated to collect multiple demonstrations.

\subsection{Discussion}
\label{sec:discussion}
\paragraph{Guiding Trajectory Distribution with LM Components.} To better understand how the LM labeler and policy shape the distribution of trajectories, we consider how this distribution evolves over the course of multiple iterations.  Let $p_k(\tau)$ be the distribution over trajectories and $p_k(g)$ be the distribution over instructions, after $k$ iterations. For $k > 0$:
\begin{align}
    p_k(\tau) &= \sum_{g'} p_\text{agent}(\tau \mid g') \cdot p_{k-1}(g')\\
    p_{k-1}(g') &= \sum_{\tau'} p_\text{label}(\tau' | g') \cdot p_{k-1}(\tau').
\end{align}

Combining these, we obtain:
\begin{align}
    p_k(\tau) = \sum_{\tau', g'} p_{k-1}(\tau') \cdot \underbrace{ p_{\text{label}}(g' \mid \tau') \cdot p_{\text{agent}}(\tau \mid g')}_{\text{environment and LM constraints}}.
\end{align}
Thus, we shape the distribution of trajectories from the previous marginal $p_{k-1}$ based on the criteria that they can be assigned a concrete string $g'$, and are executable in the environment. These soft constraints work together to ensure that (1) trajectories can be described in terms of some feasible instruction in the environment, and (2) the trajectories themselves correspond to valid environment dynamics.

\paragraph{Connection to Hindsight Experience Replay.} Hindsight Experience Replay (HER, \citealp{andrychowicz2017hindsight}) is a popular approach for training language conditioned policies. Given some goal $g$, HER converts an unsuccessful trajectory $\tau$ into positive examples by replacing $g$ with some \emph{hindsight goal} $g'$. That is, HER uses a \emph{relabeling function} to map $\tau$ to a new goal $g'$, resulting in a positive demonstration $(g', \tau)$, that is used to update the policy.

Since the original implementation of HER considers settings where the goal space is the raw environment observation space, applying HER to natural language instruction-following requires access to a learnt relabeling function to map observations to language instructions. Such relabeling functions typically map only the \emph{final} observation $o_T$ to the instruction via pre-trained captioning models \cite{xiao2022robotic, cideron2020higher, sumers2023distilling} that operate on trajectories from \emph{trained} agents. In $\ours{}$, we use the full trajectory for relabeling and use an iterative relabeling procedure to reduce noise from zero-shot components.  

\section{Inference} We use synthetic demonstrations from $\ours{}$ to adapt LM agents via retrieval augmented generation, and leave finetuning for future work. 
Concretely, given a test instruction $g_\text{test}$, we retrieve top-$k$ most relevant demonstrations in the demonstration set $\mathcal{M}$, pre-pending these to the context window of our agent as in-context examples. More concretely, we use dual encoder retrieval, similar to \citet{lee2019latent}, using a T5-XXL \citep{raffel2020exploring} embedding model. We first compute a vector embedding $f_\theta(g)$ for each instruction $g \in \mathcal{M}$, and then find the top-$k$ demonstrations based on scores $f_\theta(g)^{\top} f_\theta(g_{\text{test}})$. More details can be found in Appendix~\ref{sec:other_impl_details}.

\section{Datasets}

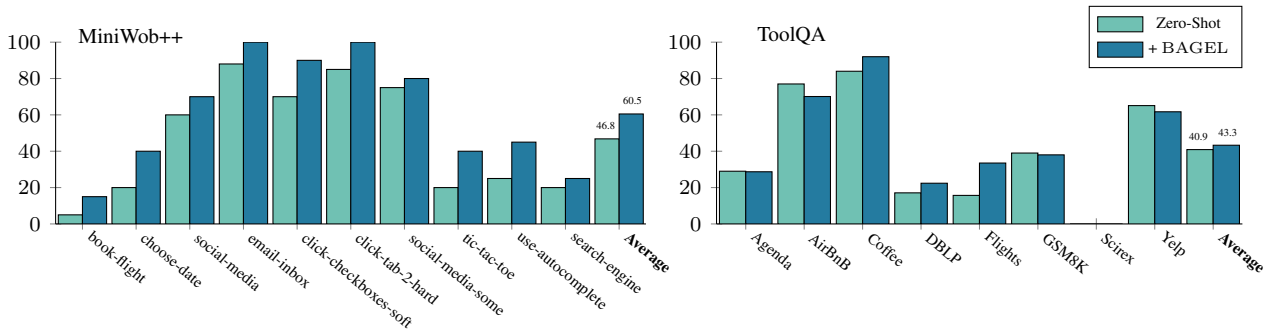
\begin{figure*}[t!]
    \begin{tikzpicture}
        \node[style={font=\fontsize{8}{10}\selectfont}] at (1,2.5) {MiniWob++};
    \begin{axis}[
        width=0.55\textwidth,
        height=4.0cm,
        ybar,
        ymin=0,
        ymax=100,
        xmin=-0.5,
        xmax=10.5,
        bar width=0.45,
        ybar=0pt,
        xtick=data,
        font=\small,
        xticklabel style={font=\fontsize{6}{8}\selectfont,yshift=0.0ex,anchor=west,rotate=320},
        xticklabels={book-flight,
        choose-date,
        social-media,
        email-inbox,
        click-checkboxes-soft,
        click-tab-2-hard,
        social-media-some,
        tic-tac-toe,
        use-autocomplete,
        search-engine,
        \textbf{Average}},
        yticklabel style={font=\fontsize{8}{10}\selectfont},
        ytick={0, 20, 40, 60, 80, 100},
        legend columns=1,
        axis lines={left},
        axis line style={-{}},
        ]
        \node[above, style={font=\fontsize{4}{8}\selectfont} ] at (axis cs:9.75,46.8) {46.8};
        \node[above, style={font=\fontsize{4}{8}\selectfont} ] at (axis cs:10.25,60.5) {60.5};
        \addplot[fill=dgreen] coordinates {
    (0, 5)
    (1, 20)
    (2, 60)
    (3, 88)
    (4, 70)
    (5, 85)
    (6, 75)
    (7, 20)
    (8, 25)
    (9, 20)
    (10, 46.8)
        };
        \addplot[fill=dblue] coordinates {
(0, 15)
(1,40)
(2,70)
(3,100)
(4,90)
(5,100)
(6,80)
(7,40)
(8,45)
(9,25)
(10,60.5)
        };
    \end{axis}
    \hspace{250pt}
    \node[style={font=\fontsize{8}{10}\selectfont}] at (1,2.5) {ToolQA};
    \begin{axis}[
        width=0.5\textwidth,
        height=4.0cm,
        ybar,
        ymin=0,
        ymax=100,
        xmin=-0.5,
        xmax=8.5,
        bar width=0.45,
        ybar=0pt,
		xlabel near ticks,
		ylabel near ticks,
        xtick=data,
        font=\small,
        xticklabel style={font=\fontsize{6}{8}\selectfont,yshift=0.0ex,anchor=west,rotate=320},
        xticklabels={
        Agenda,
        AirBnB,
        Coffee,
        DBLP,
        Flights,
        GSM8K,
        Scirex,
        Yelp,  
        \textbf{Average}},
        yticklabel style={font=\fontsize{8}{10}\selectfont},
        ytick={0, 20, 40, 60, 80, 100},
        legend image code/.code={
                    \draw[#1] (0cm,-0.1cm) rectangle (0.6cm,0.1cm);
                },
        legend style={at={(1,1.2)}, anchor=north east, font=\tiny},
        axis lines={left},
        axis line style={-{}},
        ]

        \node[above, style={font=\fontsize{4}{8}\selectfont} ] at (axis cs:7.75,40.7) {40.9};
        \node[above, style={font=\fontsize{4}{8}\selectfont} ] at (axis cs:8.25,43.2) {43.3};
        \addplot[fill=dgreen] coordinates {
(0, 29.0)
(1, 77.0)
(2, 84.0)
(3, 17.1)
(4, 15.7)
(5, 39.0)
(6, 0.0)
(7, 65.1)
(8, 40.9)
};
\addplot[fill=dblue] coordinates {
(0, 28.7)
(1, 70.1)
(2, 92.0)
(3, 22.4)
(4, 33.5)
(5, 38.0)
(6, 0.0)
(7, 61.7)
(8, 43.3)
};

\legend{Zero-Shot,+ $\ours{}$}
\end{axis}
\end{tikzpicture}

\caption{Results across MiniWoB++ and ToolQA, broken down by domain. We compare using demonstrations obtained via $\ours{}$ (\textcolor{dblue}{blue}) with a zero-shot ReAct baseline (\textcolor{dgreen}{\textbf{green}}) with no synthetic demonstrations. For MiniWob++, we use the Trajectory-First variant for exploration, and for ToolQA, we use Instruction-First. We report mean reward for MiniWob++ and F1 score for ToolQA. Overall, using $\ours{}$ demonstrations leads to improvements on both datasets.}
\vspace{-2mm}
\label{fig:main-results}
\end{figure*}

Our experiments are based on two environments, MiniWoB++ \citep{shi2017world, liu2018reinforcement} and ToolQA \citep{zhuang2023toolqa}.

\subsection{MiniWoB++}
MiniWoB++ is a collection of tasks consisting of web interfaces with a shared action space of mouse and keyboard actions. In our setup, actions are specified in natural language (\textit{Type Bob in the name text box}, \textit{Click on the datepicker}, \textit{Clear text on Destination}). The low-level controller that maps action strings into a Selenium API call is implemented via a separate zero-shot prompted LM (see Appendix~\ref{sec:lm_actions_to_api} for details). Each task consists of a script to generate variations of the task with a templated instruction, where each variation is controlled via a random seed.

\paragraph{Evaluation.} We follow \citet{shaw2023pixels} for evaluating agents on MiniWoB++, by mapping the raw MiniWoB++ reward from [-1, 1] to [0, 1]. For each web interface, we report the mean score over 50 random seeds. Starting with the set of 55 MiniWoB++ tasks used in prior work on applying LM agents to this domain \citep{gur2023understanding, kim2023language, sun2023adaplanner}, we evaluate on the hardest 10 tasks where the zero-shot agent has an average reward of less than 0.95, to perform a more targeted evaluation of $\ours{}$ to domains that are hard for zero-shot agents.

\subsection{ToolQA}
ToolQA is a tool augmented question-answering environment over 8 domains, where questions can be answered by chaining calls to multiple tools including text retrievers, databases, SQL interpreter, calculator etc. Each tool can be called according to a set of pre-defined methods (see Appendix~\ref{sec:toolqa_prompts} for the full action space for the policy and corresponding tool methods). The observation space is the string output from the most recent tool call (the first observation is hard-coded as a ``System prompt''). Each action corresponds to a specific tool call expressed in language (\textit{Load the Airbnb Database}, \textit{Calculate 3+7}), and the low-level controller is implemented by post-processing strings into tool methods. The episode terminates when the policy chooses the \textit{Finish with Answer} action e.g. \textit{Finish with Answer: 300}, where \textit{300} is taken as the predicted answer.

\paragraph{Evaluation.}  Following prior work on question-answering~\cite{rajpurkar-etal-2016-squad,rajpurkar-etal-2018-know,joshi-etal-2017-triviaqa}, we compute the F1 score of the final (free-form) model output from the \textit{Finish with Answer} tool call against ground-truth answers.

\section{Experimental Setup}

\subsection{Baselines and Ablations}
\paragraph{Zero-shot.} As our first baseline, we use the zero-shot policy $\pi_{\text{base}}$ directly at test time. 

\paragraph{Non-iterative Ablations.}
Similar in spirit to \citet{sumers2023distilling}, in \emph{$\ours$ (trajectory-first, no itrs)}, explored trajectories $\tau^0$ are labeled using $p_{\text{label}}$ and resulting demonstrations (g, $\tau^0$) are included in $\mathcal{M}$ if the score $s(g, \tau) = 1$. Similarly, in \emph{$\ours$ (instruction first, no itrs)}, synthetic instructions sampled from the instruction generator (see \S\ref{sec:model-components}) are converted into trajectories using $p_\text{agent}$, and the resulting demonstration ($g^0$, $\tau$) is added to $\mathcal{M}$, if $s(g^0, \tau) = 1$. This baseline captures a simple way to use LMs to construct synthetic demonstrations via a sample-then-filter approach: prompt an LM to generate possible instructions given the first observation from the environment, create trajectories based on these, and filter based on another criterion.  In general, we expect exploration using the instruction generator to work poorly in settings where the LM cannot predict potential instructions from just the first observation (e.g. it might hard to generate candidate instructions solely from the landing page of the website without further interaction).

\subsection{Implementation Details}
\label{sec:impl_details}
We evaluate all baselines and variants of $\ours{}$ on MiniWoB++ and ToolQA. For MiniWoB++, we start with sampling 60 trajectories in the exploration phase for trajectory-first variants of $\ours$, and sample 60 synthetic goals for instruction-first variants. For ToolQA, we sample 200 trajectories for $\ours{}$ (trajectory-first), and 200 synthetic goals for $\ours{}$ (instruction-first).

We use an instruction tuned PaLM-2 \citep{anil2023palm} as the base LM for all our experiments, and sample with a fixed temperature of 1.0. We set the max episode length $T$ to 15 for all datasets and models. We also set $T_\text{iter}$ to 5, when performing multiple iterations in $\ours{}$ \footnote{While tuning $T_\text{iter}$ on a dev set may lead to better results, we choose use a fixed value to remain truly zero-shot.}.

In addition to using ReAct prompting, we use a simple ``re-sampling'' procedure  to recover from issuing syntactically incorrect actions---if an action causes the environment to return an Exception (such as incorrectly invoking a tool, or typing on an element that cannot be typed on), we sample another action from the agent with the Exception message appended to its context. We keep re-sampling until it chooses a syntactically correct action, or terminate the episode if the agent is unable to fix an erroneous action in $m=5$ steps.

\section{Main Results}

Figure~\ref{fig:main-results} compares the zero-shot baseline with agents augmented with $\ours{}$ demonstrations. We find that using synthetic demonstrations as in-context exemplars, retrieved based on instruction relevance, lead to significant boosts in performance compared to the zero-shot agent. For the best variant of $\ours{}$, we find improvements of over 13\% points on MiniWoB++, and over 2\% on ToolQA. For MiniWoB++, our improvements are particularly strong (20\% absolute) on \emph{choose-date}, \emph{tic-tac-toe}, and \emph{use-autocomplete}. Solving these tasks successfully requires learning environment dynamics (e.g. Figure~\ref{fig:overview_fig}) which is enabled by BAGEL demonstrations. We isolate the source of these improvements from synthetic in-context exemplars in \S\ref{sec:synth_dem_analysis}. Furthermore, trajectory-first exploration significantly outperforms instruction-first on MiniWoB++, which we hypothesize is due to the LM prior being misaligned with the distribution over possible instructions on MiniWoB++.

Finally, Table~\ref{tab:mult_labels} shows that iterative re-labeling \emph{always} improves performance over non-iterative baselines. Multiple iterations of round trips improves average reward by 4-8\% on MiniWoB++ and 1.3-4.5\% on ToolQA.

\begin{table}[]
\centering
\small
\renewcommand{\arraystretch}{1.2}
\begin{tabular}{@{}lccccc@{}}\toprule
\multirow{2}{*}{Dataset} & \multirow{2}{*}{Zero-Shot} & \multicolumn{2}{c}{instruction-first} & \multicolumn{2}{c}{trajectory-first} \\
\cmidrule(lr){3-4} \cmidrule(lr){5-6} 
& & No-itrs & Full & No-itrs & Full \\ \midrule
MiniWoB++ & 46.8 & 52.0 & 56.0 & 53.0 & \textbf{61.0} \\
ToolQA & 40.9 & 38.8 & \textbf{43.3} & 40.9 & 42.2 \\ \bottomrule
\end{tabular}
\caption{Ablations showing the effect of multiple rounds of re-labeling in $\ours{}$. Multiple iterations improve performance for both instruction-first and trajectory-first variants.}
\label{tab:mult_labels}
\end{table}

\section{Analysis}

To understand how $\ours{}$ demonstrations improve agent performance, we first look at confounders from in-context learning (\S\ref{sec:synth_dem_analysis}), and then study the impact of synthetic demonstrations on execution failures (\S\ref{sec:error_recovery}). Next, we analyze the correctness (\S\ref{sec:human_filtering}) and diversity (\S\ref{sec:qualitative_analysis}) of $\ours{}$'s demonstrations to identify areas for further improvements. 

\subsection{In-context Learning with Synthetic Demonstrations} 
\label{sec:synth_dem_analysis}
In-context exemplars can provide a range of useful learning signal to LM agents, ranging from simply providing examples of valid action trajectories or relevant natural language instructions in isolation, to providing rich information about the conditional $p(\tau \mid g)$ (how to map relevant instructions into action sequences). Indeed, for some text classification tasks, \citet{min2022rethinking} find that improvements from in-context learning may be explained in terms of the former i.e.  examples of the label space and input text. To better understand how synthetic demonstrations help in our setting, we report results from two ablations. First, we provide the model with randomly chosen demonstrations instead of using the retriever (\textbf{Random}). Next, we shuffle demonstrations so that trajectories are paired with randomly chosen instruction within the set of retrieved examples (\textbf{Shuffled}). 

\paragraph{Results.} Table~\ref{tab:toolqa_shuffled_dems} reports results of these ablations. First, \textbf{Shuffled} improves performance over the zero-shot baseline, suggesting that some of the improvements come from providing examples of valid action trajectories in the domain in line with findings in~\citet{min2022rethinking}. \textbf{Ours} records a further improvement of 0.8\% over \textbf{Shuffled},  which suggests that the agent is able to use signal about the conditional to improve decision making.

\begin{table}[t!]
\centering
\small
\renewcommand{\arraystretch}{1.2}
\begin{tabular}{@{}lcccc@{}} 
\toprule
\textbf{Method} & \textbf{Accuracy} \\ \midrule
Zero-shot & 40.9 \\
Random & 38.0  \\
Shuffled & 41.4 \\
Ours & \textbf{42.2} \\ \bottomrule
\end{tabular}
\caption{Ablations showing the effect of various sources of information in synthetic demonstrations to agent performance.}
\label{tab:toolqa_shuffled_dems}
\end{table}

\begin{table}[t!]
\centering
\small
\begin{tabular}{@{}lcc@{}}
\toprule
\textbf{Task} & \textbf{Zero-Shot ($\downarrow$)} & \textbf{+BAGEL ($\downarrow$)} \\
\midrule
choose-date & 1.3 & \textbf{0.1} \\
book-flight & 3.0 & \textbf{0.6} \\
ToolQA (average) & 3.0 & \textbf{1.9} \\  \bottomrule
\end{tabular}
\caption{Average number of execution failures  for tasks in MiniWoB++ and ToolQA. We find that using synthetic demonstrations reduces execution failures.}
\label{tab:replan_steps}
\end{table}

\subsection{Synthetic demonstrations reduce execution failures}\label{sec:error_recovery}
As mentioned in \S\ref{sec:impl_details}, in our implementation, LM agents recover from execution failures using a  re-sampling procedure---when the agent generates an invalid action (such as attempting to Type on a checkbox element or calling a tool with incorrect syntax), we re-prompt it with the error message produced by the environment, until it produces a valid action. Of course, such re-sampling can be costly at inference time due to multiple calls to the LM. Table~\ref{tab:replan_steps} reports the average execution failures for tasks with re-sampling on MiniWoB++ and ToolQA. We note a considerable reduction in average re-sampling with $\ours{}$, due to a better understanding of environment dynamics, in turn leading to faster inference.

 \subsection{Correctness of Synthetic Demonstrations}\label{sec:human_filtering}
 One way to identify the scope for improvements in our method is to manually verify the correctness of demonstrations. We filter demonstrations which, upon execution, do not achieve the corresponding instruction. 
 Using these filtered demonstrations improves performance further by 7\% absolute on all 10 tasks from MiniWoB++.

\subsection{Diversity of Synthetic Demonstrations}
\label{sec:qualitative_analysis}
\begin{figure*}[t]
\begin{tikzpicture}
\node[style={font=\fontsize{8}{10}\selectfont}] at (1,2.5) {social-media};
    \begin{axis}[
        width=0.25\linewidth,
        height=4.0cm,
        ybar,
        ymin=0,
        ymax=50,
        xmin=-1,
        xmax=5,
        bar width=2.0mm,
        ylabel=\% Demonstrations,
        xtick=data,
        font=\small,
        xticklabel style={font=\fontsize{6}{8}\selectfont,xshift=-0.5ex,yshift=0.5ex,anchor=west,rotate=315},
        xticklabels={
Star,
Reply,
Forward,
Trash,
Delete,
        },
        yticklabel style={font=\fontsize{8}{10}\selectfont},
        ytick={0,10,20,30,40,50},
        axis lines={left},
        axis line style={-{}},
        ]
            	
        \addplot[fill=g-yellow] coordinates {
(0, 43.7)
(1, 33.6)
(2, 16.0)
(3, 3.4)
(4, 3.4)
        };
\end{axis}
\hspace{100pt}
\node[style={font=\fontsize{8}{10}\selectfont}] at (1.1,2.5) {email-inbox};
    \begin{axis}[
        width=0.35\linewidth,
        height=4.0cm,
        ybar,
        ymin=0,
        ymax=50,
        xmin=-1,
        xmax=10,
        bar width=2.0mm,
        xtick=data,
        font=\small,
        xticklabel style={font=\fontsize{6}{8}\selectfont,xshift=-0.5ex,yshift=0.5ex,anchor=west,rotate=315},
        xticklabels={
More,
Like,
Retweet,
Reply,
Copy,
Follow,
Mute,
Share,
Embed,
Block,
        },
        yticklabel style={font=\fontsize{8}{10}\selectfont},
        ytick={0,10,20,30,40,50},
        axis lines={left},
        axis line style={-{}},
        ]
            	
        \addplot[fill=g-yellow] coordinates {
(0, 28.6)
(1, 14.3)
(2, 14.3)
(3, 11.4)
(4, 8.6)
(5, 5.7)
(6, 5.7)
(7, 5.7)
(8, 2.9)
(9, 2.9)
        };
\end{axis}
\hspace{150pt}
\node[style={font=\fontsize{8}{10}\selectfont}] at (0.9,2.5) {ToolQA};
    \begin{axis}[
        width=0.45\linewidth,
        height=4.0cm,
        ybar,
        ymin=0,
        ymax=50,
        xmin=-1,
        xmax=10,
        bar width=2.0mm,
        xtick=data,
        font=\small,
        xticklabel style={font=\fontsize{6}{8}\selectfont,xshift=-0.5ex,yshift=0.5ex,anchor=west,rotate=315},
        xticklabels={
\{Graph\},
\{SQL\text{,} Database\},
\{Calculator\},
\{Scirex\},
\{Agenda\},
\{Database\},
\{Python\},
\{Python\text{,} Graph\},
\{Python\text{,} Scirex\},
\{Scirex\text{,} Graph\},
        },
        yticklabel style={font=\fontsize{8}{10}\selectfont},
        ytick={0,10,20,30,40,50},
        axis lines={left},
        axis line style={-{}},
        ]
            	
        \addplot[fill=g-yellow] coordinates {
(0, 36.5)
(1, 15.1)
(2, 13.5)
(3, 10.3)
(4, 8.7)
(5, 7.9)
(6, 4.8)
(7, 1.6)
(8, 0.8)
(9, 0.8)
        };
\end{axis}
\end{tikzpicture}
\caption{Distribution of demonstrations over semantic categories for MiniWob++ environments, \textit{social-media} and \texttt{email-inbox}, and ToolQA. While $\ours{}$ prefers certain modes, overall we find that these demonstrations cover a diverse range of actions.} 
\label{fig:diversity}
\end{figure*}
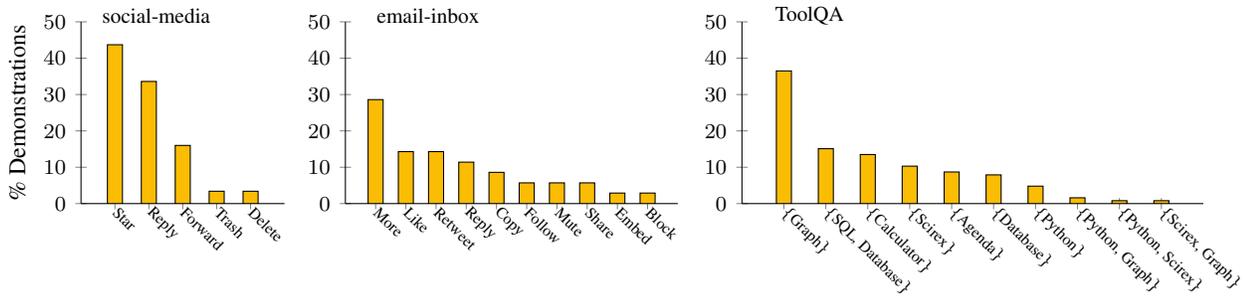

To better understand the distribution of synthetic demonstrations, we manually bucket demonstrations for \textit{social-media} and \textit{email-inbox} into semantic clusters--- for \textit{social-media} these clusters include \{Retweet, Like, Share, ...\} and for \textit{email-inbox} we have clusters such as \{Forward, Delete, Star, Reply, ...\}. For ToolQA, we cluster demonstrations based on the set of tools invoked in the demonstration. We plot the number of demonstrations in each cluster in Figure~\ref{fig:diversity}. We note that while this distribution tends to be skewed towards specific modes (e.g. \{graph\} for ToolQA, \{Star\} for email-inbox), there exists a long tail that covers a broad range of possible use cases in the environment.   Nevertheless, improving diversity during exploration remains a failure mode for $\ours$ which we expand on next. Finally, we provide some examples of $\ours{}$ demonstrations in Table~\ref{tab:qual_examples}, along with their corresponding semantic category.

\subsection{Error Analysis}
\begin{table*}[h]
\centering
\small
\renewcommand{\arraystretch}{1.2}
\begin{tabularx}{\textwidth}{XX}
\toprule
\textbf{Instruction} & \textbf{Trajectory} \\ \midrule
\multicolumn{2}{c}{MiniWoB++} \\ \hdashline
Find the email by Trixi and reply to them with the text "Maecenas eu massa" \textbf{\{Reply\}} & 
Move Mouse to Trixi $\rightarrow$ Click on an email-thread $\rightarrow$ Click on the reply button $\rightarrow$ Type 'Maecenas eu massa' on the textarea with id 'reply-text' $\rightarrow$
Click on the span with id 'send-reply' \\
Find the email by Darcy and forward it to Dionis  \textbf{\{Forward\}}& 
Click on Darcy, the sender of an email thread. $\rightarrow$ Click on 'forward' button $\rightarrow$ Type Dionis on the to field $\rightarrow$ Click on the 'send' button \\
Retweet Gallegos's post  \textbf{\{Retweet\}} & Move Mouse to Pretium,. Ullamcorper. $\rightarrow$ Click on retweet element with id 101 \\ 
Like tweet by @leonie and share tweet by @livia \textbf{\{Like, Share\}} &  Click on the like element with ID 41. $\rightarrow$ Click on share-113 \\ \midrule

\multicolumn{2}{c}{ToolQA} \\ \hdashline
What are David's plans this weekend? \textbf{\{RetrieveAgenda\}} & Retrieve passages related to David's plans this weekend $\rightarrow$ Finish with answer: On the evening of September 16th, 2022, David will be attending a Blind Date Night event at The Press Lounge. \\
Who is affiliated with both nicolas christin and giulia fanti? \textbf{\{Python, Graph\}} & Load DBLP $\rightarrow$ Check neighbours of node Giulia Fanti in graph AuthorNet $\rightarrow$ Check neighbours of node Nicolas Christin in graph AuthorNet $\rightarrow$ Evaluate python code: \textit{list1=['Wanzheng Zhu', 'Rohan Bansal', 'Zachary Weinberg', 'Nicolas Christin', 'Suma Bhat', 'Hongyu Gong']; list2=['Wanzheng Zhu', 'Rohan Bansal', 'Zachary Weinberg', 'Suma Bhat', 'Hongyu Gong', 'Giulia Fanti']; ans=set(list1) \& set(list2)} $\rightarrow$ Finish with answer: \{'Hongyu Gong', 'Rohan Bansal', 'Wanzheng Zhu', 'Zachary Weinberg', 'Suma Bhat'\} \\
What are the top 5 airbnb options with price $<$ 900, availability $>$ 260 and at least 40 reviews \textbf{\{Database, SQL\}} & Load database airbnb $\rightarrow$ Filter database according to price $<$ 900, availability 365 $>$ 260, number of reviews $>$ 40 $\rightarrow$ Interpret SQLite query: 
\textit{SELECT \* FROM airbnb\_data ORDER BY number\_of\_reviews DESC LIMIT 5} $\rightarrow$
Finish with answer: ['High-end doorman bldg in the LES', 'THE BEST DEAL ON THE HUDSON RIVER!!', 'Heart of Williamsburg, Brooklyn!', 'Beautiful \& Tranquil Oasis in a Great Location', 'Sunny/Cozy 1BD'] \\
What are the different approaches for computing graph similarity? \textbf{\{RetrieveSciRex\}} &  Retrieve passages from ML papers related to graph similarity $\rightarrow$ Finish with answer: The different approaches to computing graph similarity are graph kernels, graph features and graph convolutional neural networks (CNNs). \\ \bottomrule
\end{tabularx}
\caption{Example demonstrations obtained via $\ours{}$ for MiniWoB++ (top) and ToolQA (bottom). We also provide the semantic category for these demonstrations, and report the distribution of these categories in Figure~\ref{fig:diversity}.}
\label{tab:qual_examples}
\end{table*}

We conclude with a discussion of failure modes of our aproach using the domains \textit{book-flight}, \textit{search-engine}, and \textit{SciRex} as case studies.
\paragraph{Handling Long-Horizon Planning.} We note that \textit{book-flight} is the most complex environment in MiniWoB++, with longer trajectories of lengths 8-20, and the zero-shot policy performs poorly on this environment (average reward of 5\%). While using BAGEL demonstrations improves this to 15\%, we hypothesize that further improvements would require better handling of long range plans, such as with hierarchical planning \citep{sodhi2023heap, jiang2019language}.

\paragraph{Improving Diversity.} We hypothesize that improving diversity among seed trajectories would lead to further improvements across the board. For instance, for \textit{book-flight}, all $\ours{}$ demonstrations correspond to booking flights in December, while the test distribution is more uniform. 

\paragraph{Reducing Mismatch with Test Instructions.}  On \textit{SciRex}, all models fail to produce even a single correct answer. Here, we find that in the absence of any knowledge about user instructions at test-time, $\ours{}$ demonstrations tend to create questions with more descriptive answers and trajectories with generic queries (See Table~\ref{tab:qual_examples} for an example) while test instructions requires retrieving specific numbers from scientific documents by querying for specific topics. Similarly, on \textit{search-engine}, we note a modest improvement of only 5\%. Here, we find that while $\ours{}$ demonstrations cover a variety of instructions like \textit{Search for cat and navigate to the third page of search results, Search for cars, then visit the second search result}, the model fails on test instructions like  \textit{Enter [term] then find and click the 9th search result} that requires keeping track of the number of search results per page, and navigating to the correct page. While our goal is to build fully unsupervised agents, methods that use sparse information about test-time instructions could help drive performance further.

\section{Related Work}
\paragraph{Instruction-Following Digital Agents.} 
Building agents that navigate the digital world is a long standing goal of AI and language understanding \citep{ allen2007agents, branavan2009reinforcement}.
However, most prior work relies on expert demonstrations \citep{liu2018reinforcement, humphreys2022data,furuta2023multimodal} with an appropriately shaped reward \citep{branavan2009reinforcement, liu2018reinforcement}. Here, we assume no access to demonstrations or a reward function, and use pre-trained components to bootstrap synthetic demonstrations.

\paragraph{LMs for Decision Making.} Pre-trained LMs are increasingly being used for sequential tasks such as robotic manipulation \citep{ahn2022can, liang2023code}, instruction-following \citep{yao2022react, kim2023language, sun2023adaplanner, lu2024weblinx}, and tool-use \citep{parisi2022talm}. While some of these approaches finetune LMs based on human demonstrations \citep{nakano2021webgpt}, others use human demonstrations in their prompt for in-context learning and adaptation \citep{yao2022react, kim2023language, sun2023adaplanner}. We use no human supervision or reward and adapt LM agents purely using synthetic demonstrations. Another line of work uses LM priors in reinforcement learning to improve exploration \citep{mu2022improving, du2023guiding}, deal with large action spaces \citep{yao2020keep}, or as proxy reward functions \citep{kwon2023reward}. In the same tradition, $\ours{}$  bootstraps a learning signal in the form of synthetic demonstrations by combining several LM components but without using RL.

\paragraph{Self-training for Language Models.} A recent line of work uses LM-generated data for finetuning the same LM, in settings where external verifiers may be used to filter generated data \citep{singh2023beyond, gulcehre2023reinforced}. While we also use data generated from an LM for adaptation, unlike these approaches, environment interactions form a critical part of the learning signal and we also do not use external verifiers for filtering data.

\section{Conclusion}
There is a growing interesting in grounding LMs to the real world, by building helpful assistants that execute open-ended instructions in digital environments. The complexity of such sequential tasks makes collecting expert demonstrations tedious, and so, further progress towards building such agents requires new methods for bootstrapping a learning signal with minimal human supervision.
To this end, we introduce $\ours{}$, a method for constructing synthetic demonstrations for instruction following agents. These demonstrations are constructed by iteratively relabeling an initial seed set of trajectories or instructions, where both relabeling and exploration is driven by a language model. Experiments on two different domains show that using $\ours{}$ demonstrations as in-context exemplars leads to considerable improvements ranging from 2-13\%, as well as significant reductions in execution failures.

\section*{Impact Statement}
In this paper, we evaluated models only in offline environments. Responsibly deploying models online carries potential risks, and it would be important to verify and constrain model behaviour to not cause harm (e.g. violating terms of service). Further research related to secure model deployment should take into account problems such as spam detection, privacy preservation, etc.

\section*{Acknowledgements}
SM was partly funded by a gift from Apple Inc. CM is a fellow in the CIFAR Learning in Machines and Brains program. We thank David Gaddy, Anna Goldie, Luke Vilnis, Tianze Shi, Jonathan Berant, Kristina Toutanova, Raphael Hoffman, and members of Google DeepMind and the Stanford NLP Group for helpful discussions and comments. 

\bibliography{icml2024-ref}

\begin{thebibliography}{41}
\providecommand{\natexlab}[1]{#1}
\providecommand{\url}[1]{\texttt{#1}}
\expandafter\ifx\csname urlstyle\endcsname\relax
  \providecommand{\doi}[1]{doi: #1}\else
  \providecommand{\doi}{doi: \begingroup \urlstyle{rm}\Url}\fi

\bibitem[Ahn et~al.(2022)Ahn, Brohan, Brown, Chebotar, Cortes, David, Finn, Fu,
  Gopalakrishnan, Hausman, et~al.]{ahn2022can}
Ahn, M., Brohan, A., Brown, N., Chebotar, Y., Cortes, O., David, B., Finn, C.,
  Fu, C., Gopalakrishnan, K., Hausman, K., et~al.
\newblock Do as i can, not as i say: Grounding language in robotic affordances.
\newblock \emph{arXiv preprint arXiv:2204.01691}, 2022.

\bibitem[Allen et~al.(2007)Allen, Chambers, Ferguson, Galescu, Jung, Swift, and
  Taysom]{allen2007agents}
Allen, J., Chambers, N., Ferguson, G., Galescu, L., Jung, H., Swift, M., and
  Taysom, W.
\newblock Plow: a collaborative task learning agent.
\newblock In \emph{Proceedings of the 22nd National Conference on Artificial
  Intelligence - Volume 2}, AAAI'07, pp.\  1514–1519. AAAI Press, 2007.
\newblock ISBN 9781577353232.

\bibitem[Andrychowicz et~al.(2017)Andrychowicz, Wolski, Ray, Schneider, Fong,
  Welinder, McGrew, Tobin, Pieter~Abbeel, and
  Zaremba]{andrychowicz2017hindsight}
Andrychowicz, M., Wolski, F., Ray, A., Schneider, J., Fong, R., Welinder, P.,
  McGrew, B., Tobin, J., Pieter~Abbeel, O., and Zaremba, W.
\newblock Hindsight experience replay.
\newblock \emph{Advances in neural information processing systems}, 30, 2017.

\bibitem[Anil et~al.(2023)Anil, Dai, Firat, Johnson, Lepikhin, Passos, Shakeri,
  Taropa, Bailey, Chen, et~al.]{anil2023palm}
Anil, R., Dai, A.~M., Firat, O., Johnson, M., Lepikhin, D., Passos, A.,
  Shakeri, S., Taropa, E., Bailey, P., Chen, Z., et~al.
\newblock Palm 2 technical report.
\newblock \emph{arXiv preprint arXiv:2305.10403}, 2023.

\bibitem[Branavan et~al.(2009)Branavan, Chen, Zettlemoyer, and
  Barzilay]{branavan2009reinforcement}
Branavan, S., Chen, H., Zettlemoyer, L., and Barzilay, R.
\newblock Reinforcement learning for mapping instructions to actions.
\newblock In Su, K.-Y., Su, J., Wiebe, J., and Li, H. (eds.), \emph{Proceedings
  of the Joint Conference of the 47th Annual Meeting of the {ACL} and the 4th
  International Joint Conference on Natural Language Processing of the
  {AFNLP}}, pp.\  82--90, Suntec, Singapore, August 2009. Association for
  Computational Linguistics.
\newblock URL \url{https://aclanthology.org/P09-1010}.

\bibitem[Chaplot et~al.(2018)Chaplot, Sathyendra, Pasumarthi, Rajagopal, and
  Salakhutdinov]{chaplot2018gated}
Chaplot, D.~S., Sathyendra, K.~M., Pasumarthi, R.~K., Rajagopal, D., and
  Salakhutdinov, R.
\newblock Gated-attention architectures for task-oriented language grounding.
\newblock In \emph{Proceedings of the AAAI Conference on Artificial
  Intelligence}, volume~32, 2018.

\bibitem[Cideron et~al.(2020)Cideron, Seurin, Strub, and
  Pietquin]{cideron2020higher}
Cideron, G., Seurin, M., Strub, F., and Pietquin, O.
\newblock Higher: Improving instruction following with hindsight generation for
  experience replay.
\newblock In \emph{2020 IEEE Symposium Series on Computational Intelligence
  (SSCI)}, pp.\  225--232. IEEE, 2020.

\bibitem[Du et~al.(2023)Du, Watkins, Wang, Colas, Darrell, Abbeel, Gupta, and
  Andreas]{du2023guiding}
Du, Y., Watkins, O., Wang, Z., Colas, C., Darrell, T., Abbeel, P., Gupta, A.,
  and Andreas, J.
\newblock Guiding pretraining in reinforcement learning with large language
  models.
\newblock \emph{arXiv preprint arXiv:2302.06692}, 2023.

\bibitem[Furuta et~al.(2023)Furuta, Nachum, Lee, Matsuo, Gu, and
  Gur]{furuta2023multimodal}
Furuta, H., Nachum, O., Lee, K.-H., Matsuo, Y., Gu, S.~S., and Gur, I.
\newblock Multimodal web navigation with instruction-finetuned foundation
  models.
\newblock \emph{arXiv preprint arXiv:2305.11854}, 2023.

\bibitem[Gulcehre et~al.(2023)Gulcehre, Paine, Srinivasan, Konyushkova, Weerts,
  Sharma, Siddhant, Ahern, Wang, Gu, et~al.]{gulcehre2023reinforced}
Gulcehre, C., Paine, T.~L., Srinivasan, S., Konyushkova, K., Weerts, L.,
  Sharma, A., Siddhant, A., Ahern, A., Wang, M., Gu, C., et~al.
\newblock Reinforced self-training (rest) for language modeling.
\newblock \emph{arXiv preprint arXiv:2308.08998}, 2023.

\bibitem[Gur et~al.(2023)Gur, Nachum, Miao, Safdari, Huang, Chowdhery, Narang,
  Fiedel, and Faust]{gur2023understanding}
Gur, I., Nachum, O., Miao, Y., Safdari, M., Huang, A., Chowdhery, A., Narang,
  S., Fiedel, N., and Faust, A.
\newblock Understanding {HTML} with large language models.
\newblock In Bouamor, H., Pino, J., and Bali, K. (eds.), \emph{Findings of the
  Association for Computational Linguistics: EMNLP 2023}, pp.\  2803--2821,
  Singapore, December 2023. Association for Computational Linguistics.
\newblock \doi{10.18653/v1/2023.findings-emnlp.185}.
\newblock URL \url{https://aclanthology.org/2023.findings-emnlp.185}.

\bibitem[Huang et~al.(2022)Huang, Abbeel, Pathak, and
  Mordatch]{huang2022language}
Huang, W., Abbeel, P., Pathak, D., and Mordatch, I.
\newblock Language models as zero-shot planners: Extracting actionable
  knowledge for embodied agents.
\newblock In \emph{International Conference on Machine Learning}, pp.\
  9118--9147. PMLR, 2022.

\bibitem[Humphreys et~al.(2022)Humphreys, Raposo, Pohlen, Thornton, Chhaparia,
  Muldal, Abramson, Georgiev, Santoro, and Lillicrap]{humphreys2022data}
Humphreys, P.~C., Raposo, D., Pohlen, T., Thornton, G., Chhaparia, R., Muldal,
  A., Abramson, J., Georgiev, P., Santoro, A., and Lillicrap, T.
\newblock A data-driven approach for learning to control computers.
\newblock In \emph{International Conference on Machine Learning}, pp.\
  9466--9482. PMLR, 2022.

\bibitem[Jiang et~al.(2019)Jiang, Gu, Murphy, and Finn]{jiang2019language}
Jiang, Y., Gu, S.~S., Murphy, K.~P., and Finn, C.
\newblock Language as an abstraction for hierarchical deep reinforcement
  learning.
\newblock \emph{Advances in Neural Information Processing Systems}, 32, 2019.

\bibitem[Joshi et~al.(2017)Joshi, Choi, Weld, and
  Zettlemoyer]{joshi-etal-2017-triviaqa}
Joshi, M., Choi, E., Weld, D., and Zettlemoyer, L.
\newblock {T}rivia{QA}: A large scale distantly supervised challenge dataset
  for reading comprehension.
\newblock In Barzilay, R. and Kan, M.-Y. (eds.), \emph{Proceedings of the 55th
  Annual Meeting of the Association for Computational Linguistics (Volume 1:
  Long Papers)}, pp.\  1601--1611, Vancouver, Canada, July 2017. Association
  for Computational Linguistics.
\newblock \doi{10.18653/v1/P17-1147}.
\newblock URL \url{https://aclanthology.org/P17-1147}.

\bibitem[Kim et~al.(2023)Kim, Baldi, and McAleer]{kim2023language}
Kim, G., Baldi, P., and McAleer, S.
\newblock Language models can solve computer tasks.
\newblock \emph{arXiv preprint arXiv:2303.17491}, 2023.

\bibitem[Kwon et~al.(2023)Kwon, Xie, Bullard, and Sadigh]{kwon2023reward}
Kwon, M., Xie, S.~M., Bullard, K., and Sadigh, D.
\newblock Reward design with language models.
\newblock In \emph{The Eleventh International Conference on Learning
  Representations}, 2023.
\newblock URL \url{https://openreview.net/forum?id=10uNUgI5Kl}.

\bibitem[Lee et~al.(2019)Lee, Chang, and Toutanova]{lee2019latent}
Lee, K., Chang, M.-W., and Toutanova, K.
\newblock Latent retrieval for weakly supervised open domain question
  answering.
\newblock In \emph{Proceedings of the 57th Annual Meeting of the Association
  for Computational Linguistics}, pp.\  6086--6096, 2019.

\bibitem[Liang et~al.(2023)Liang, Huang, Xia, Xu, Hausman, Ichter, Florence,
  and Zeng]{liang2023code}
Liang, J., Huang, W., Xia, F., Xu, P., Hausman, K., Ichter, B., Florence, P.,
  and Zeng, A.
\newblock Code as policies: Language model programs for embodied control.
\newblock In \emph{2023 IEEE International Conference on Robotics and
  Automation (ICRA)}, pp.\  9493--9500. IEEE, 2023.

\bibitem[Liu et~al.(2018)Liu, Guu, Pasupat, Shi, and
  Liang]{liu2018reinforcement}
Liu, E.~Z., Guu, K., Pasupat, P., Shi, T., and Liang, P.
\newblock Reinforcement learning on web interfaces using workflow-guided
  exploration.
\newblock In \emph{International Conference on Learning Representations}, 2018.

\bibitem[Logeswaran et~al.(2022)Logeswaran, Fu, Lee, and
  Lee]{logeswaran2022few}
Logeswaran, L., Fu, Y., Lee, M., and Lee, H.
\newblock Few-shot subgoal planning with language models.
\newblock In \emph{Proceedings of the 2022 Conference of the North American
  Chapter of the Association for Computational Linguistics: Human Language
  Technologies}, pp.\  5493--5506, 2022.

\bibitem[L{\`u} et~al.(2024)L{\`u}, Kasner, and Reddy]{lu2024weblinx}
L{\`u}, X.~H., Kasner, Z., and Reddy, S.
\newblock Weblinx: Real-world website navigation with multi-turn dialogue.
\newblock \emph{arXiv preprint arXiv:2402.05930}, 2024.

\bibitem[Min et~al.(2022)Min, Lyu, Holtzman, Artetxe, Lewis, Hajishirzi, and
  Zettlemoyer]{min2022rethinking}
Min, S., Lyu, X., Holtzman, A., Artetxe, M., Lewis, M., Hajishirzi, H., and
  Zettlemoyer, L.
\newblock Rethinking the role of demonstrations: What makes in-context learning
  work?
\newblock \emph{arXiv preprint arXiv:2202.12837}, 2022.

\bibitem[Misra et~al.(2017)Misra, Langford, and Artzi]{misra2017mapping}
Misra, D., Langford, J., and Artzi, Y.
\newblock Mapping instructions and visual observations to actions with
  reinforcement learning.
\newblock \emph{arXiv preprint arXiv:1704.08795}, 2017.

\bibitem[Mu et~al.(2022)Mu, Zhong, Raileanu, Jiang, Goodman, Rockt{\"a}schel,
  and Grefenstette]{mu2022improving}
Mu, J., Zhong, V., Raileanu, R., Jiang, M., Goodman, N., Rockt{\"a}schel, T.,
  and Grefenstette, E.
\newblock Improving intrinsic exploration with language abstractions.
\newblock \emph{Advances in Neural Information Processing Systems},
  35:\penalty0 33947--33960, 2022.

\bibitem[Nakano et~al.(2021)Nakano, Hilton, Balaji, Wu, Ouyang, Kim, Hesse,
  Jain, Kosaraju, Saunders, et~al.]{nakano2021webgpt}
Nakano, R., Hilton, J., Balaji, S., Wu, J., Ouyang, L., Kim, C., Hesse, C.,
  Jain, S., Kosaraju, V., Saunders, W., et~al.
\newblock Webgpt: Browser-assisted question-answering with human feedback.
\newblock \emph{arXiv preprint arXiv:2112.09332}, 2021.

\bibitem[Parisi et~al.(2022)Parisi, Zhao, and Fiedel]{parisi2022talm}
Parisi, A., Zhao, Y., and Fiedel, N.
\newblock Talm: Tool augmented language models.
\newblock \emph{arXiv preprint arXiv:2205.12255}, 2022.

\bibitem[Raffel et~al.(2020)Raffel, Shazeer, Roberts, Lee, Narang, Matena,
  Zhou, Li, and Liu]{raffel2020exploring}
Raffel, C., Shazeer, N., Roberts, A., Lee, K., Narang, S., Matena, M., Zhou,
  Y., Li, W., and Liu, P.~J.
\newblock Exploring the limits of transfer learning with a unified text-to-text
  transformer.
\newblock \emph{Journal of Machine Learning Research}, 21:\penalty0 1--67,
  2020.

\bibitem[Rajpurkar et~al.(2016)Rajpurkar, Zhang, Lopyrev, and
  Liang]{rajpurkar-etal-2016-squad}
Rajpurkar, P., Zhang, J., Lopyrev, K., and Liang, P.
\newblock {SQ}u{AD}: 100,000+ questions for machine comprehension of text.
\newblock In Su, J., Duh, K., and Carreras, X. (eds.), \emph{Proceedings of the
  2016 Conference on Empirical Methods in Natural Language Processing}, pp.\
  2383--2392, Austin, Texas, November 2016. Association for Computational
  Linguistics.
\newblock \doi{10.18653/v1/D16-1264}.
\newblock URL \url{https://aclanthology.org/D16-1264}.

\bibitem[Rajpurkar et~al.(2018)Rajpurkar, Jia, and
  Liang]{rajpurkar-etal-2018-know}
Rajpurkar, P., Jia, R., and Liang, P.
\newblock Know what you don{'}t know: Unanswerable questions for {SQ}u{AD}.
\newblock In Gurevych, I. and Miyao, Y. (eds.), \emph{Proceedings of the 56th
  Annual Meeting of the Association for Computational Linguistics (Volume 2:
  Short Papers)}, pp.\  784--789, Melbourne, Australia, July 2018. Association
  for Computational Linguistics.
\newblock \doi{10.18653/v1/P18-2124}.
\newblock URL \url{https://aclanthology.org/P18-2124}.

\bibitem[Shaw et~al.(2023)Shaw, Joshi, Cohan, Berant, Pasupat, Hu, Khandelwal,
  Lee, and Toutanova]{shaw2023pixels}
Shaw, P., Joshi, M., Cohan, J., Berant, J., Pasupat, P., Hu, H., Khandelwal,
  U., Lee, K., and Toutanova, K.
\newblock From pixels to ui actions: Learning to follow instructions via
  graphical user interfaces.
\newblock \emph{arXiv preprint arXiv:2306.00245}, 2023.

\bibitem[Shi et~al.(2017)Shi, Karpathy, Fan, Hernandez, and
  Liang]{shi2017world}
Shi, T., Karpathy, A., Fan, L., Hernandez, J., and Liang, P.
\newblock World of bits: An open-domain platform for web-based agents.
\newblock In \emph{International Conference on Machine Learning}, pp.\
  3135--3144. PMLR, 2017.

\bibitem[Shinn et~al.(2023)Shinn, Labash, and Gopinath]{shinn2023reflexion}
Shinn, N., Labash, B., and Gopinath, A.
\newblock Reflexion: an autonomous agent with dynamic memory and
  self-reflection.
\newblock \emph{arXiv preprint arXiv:2303.11366}, 2023.

\bibitem[Singh et~al.(2023)Singh, Co-Reyes, Agarwal, Anand, Patil, Liu,
  Harrison, Lee, Xu, Parisi, et~al.]{singh2023beyond}
Singh, A., Co-Reyes, J.~D., Agarwal, R., Anand, A., Patil, P., Liu, P.~J.,
  Harrison, J., Lee, J., Xu, K., Parisi, A., et~al.
\newblock Beyond human data: Scaling self-training for problem-solving with
  language models.
\newblock \emph{arXiv preprint arXiv:2312.06585}, 2023.

\bibitem[Sodhi et~al.(2023)Sodhi, Branavan, and McDonald]{sodhi2023heap}
Sodhi, P., Branavan, S., and McDonald, R.
\newblock Heap: Hierarchical policies for web actions using llms.
\newblock \emph{arXiv preprint arXiv:2310.03720}, 2023.

\bibitem[Sumers et~al.(2023)Sumers, Marino, Ahuja, Fergus, and
  Dasgupta]{sumers2023distilling}
Sumers, T., Marino, K., Ahuja, A., Fergus, R., and Dasgupta, I.
\newblock Distilling internet-scale vision-language models into embodied
  agents.
\newblock 2023.

\bibitem[Sun et~al.(2023)Sun, Zhuang, Kong, Dai, and Zhang]{sun2023adaplanner}
Sun, H., Zhuang, Y., Kong, L., Dai, B., and Zhang, C.
\newblock Adaplanner: Adaptive planning from feedback with language models.
\newblock \emph{arXiv preprint arXiv:2305.16653}, 2023.

\bibitem[Xiao et~al.(2022)Xiao, Chan, Sermanet, Wahid, Brohan, Hausman, Levine,
  and Tompson]{xiao2022robotic}
Xiao, T., Chan, H., Sermanet, P., Wahid, A., Brohan, A., Hausman, K., Levine,
  S., and Tompson, J.
\newblock Robotic skill acquisition via instruction augmentation with
  vision-language models.
\newblock \emph{arXiv preprint arXiv:2211.11736}, 2022.

\bibitem[Yao et~al.(2020)Yao, Rao, Hausknecht, and Narasimhan]{yao2020keep}
Yao, S., Rao, R., Hausknecht, M., and Narasimhan, K.
\newblock Keep {CALM} and explore: Language models for action generation in
  text-based games.
\newblock In Webber, B., Cohn, T., He, Y., and Liu, Y. (eds.),
  \emph{Proceedings of the 2020 Conference on Empirical Methods in Natural
  Language Processing (EMNLP)}, pp.\  8736--8754, Online, November 2020.
  Association for Computational Linguistics.
\newblock \doi{10.18653/v1/2020.emnlp-main.704}.
\newblock URL \url{https://aclanthology.org/2020.emnlp-main.704}.

\bibitem[Yao et~al.(2022)Yao, Zhao, Yu, Du, Shafran, Narasimhan, and
  Cao]{yao2022react}
Yao, S., Zhao, J., Yu, D., Du, N., Shafran, I., Narasimhan, K.~R., and Cao, Y.
\newblock React: Synergizing reasoning and acting in language models.
\newblock In \emph{The Eleventh International Conference on Learning
  Representations}, 2022.

\bibitem[Zhuang et~al.(2023)Zhuang, Yu, Wang, Sun, and Zhang]{zhuang2023toolqa}
Zhuang, Y., Yu, Y., Wang, K., Sun, H., and Zhang, C.
\newblock Toolqa: A dataset for llm question answering with external tools.
\newblock \emph{arXiv preprint arXiv:2306.13304}, 2023.

\end{thebibliography}
\bibliographystyle{icml2024}

\newpage
\appendix
\onecolumn
\section{Other Implementation Details}
\label{sec:other_impl_details}

\subsection{Retriever}
We use a T5-XXL model to embed each word in the instruction, and mean pool across word embeddings to obtain an instruction vector. Given a test-time instruction, to retrieve relevant demonstrations, we compute cosine similarities between the test instruction embedding and instruction embeddings for each demonstration in our buffer, and return the top 3 demonstrations with the highest cosine similarities. 

\subsection{Re-sampling action strings} 
When executing an action string in the environment results in an exception from the low-level controller, we pass the exception message to the LM policy, and re-sample till the model outputs a valid action, or the LM exceeds the max number of tries $m = 5$. Here is an example prompt we use for this re-sampling procedure (the prompt is appended to the LM policy):

\begin{lstlisting}[style=textFileStyle, caption=Re-sampling during Execution Failure ]
Executing Action: {error_action}... 
resulted in error: {error_message}. Think about what could have caused the error, and then choose a new action.

Thought: [[thought_pred]]

Now, output a different action based on your thought. End your output with a newline.

Action: [[action]]
\end{lstlisting}

\section{Prompts}
\label{sec:prompts}

\subsection{MiniWoB++} 
\label{sec:miniwob_prompts}

We start by presenting all prompts for MiniWoB++. The action space for MiniWob++ is:

\begin{lstlisting}[style=textFileStyle, caption=Action Space ]
- Click on *description*: This action will click on element that matches *description* e.g. Click on the red button, Click on the first result in the autocomplete
- Move Mouse to *description*: This action will hover mouse over web element that matches *description* e.g. Move mouse to the menu bar.
- Type char *char* on *description*: This action will type a single character *char* into the web element matching *description* e.g. Type char B on the first name field. Use this if you want to type in a word character by character, to view or narrow search results.
- Type *text* on *description*: This action will type *text* into the web element matching *description*. Use this to type in all the words in *text*' all at once.
- Clear text on *description*: This action will clear all previously typed text in web element matching *description*
\end{lstlisting}
\vspace{2em}
This is then directly used for various prompts as \texttt{\{inventory\_str\}}.


\begin{lstlisting}[style=textFileStyle, caption=Exploration Policy]
You are a web-agent that can interact with the given webpage by taking actions. You can take the following kinds of actions:
{inventory_str}

Your objective is to discover diverse and interesting tasks (that a human might give to an agent) by interacting with the webpage through these actions. You've executed the following actions, and observed the following webpage states (described briefly in language).

**Previous observations and actions**
{prev_observations_and_actions}

After taking these actions, you observe the current web-page HTML:
{webpage_html} 

Start by thinking about what action you should take next.
Thought: [[pred]]

Now, act by taking an action based in the inventory (or output Finish if you are done).
Action: [[pred]]
\end{lstlisting}

\begin{lstlisting}[style=textFileStyle, caption=Instruction Generator]
**Objective**
You are a web-agent that can accomplish useful tasks on a website. You are given the landing page of the website as follows:
{init_html}

To accomplish tasks, you can break it down into a sequence of sub-tasks from a task inventory:
{inventory_str}

Propose a new task that can be performed on this website. Ensure that your tasks are concrete and use features / contents of the given website. 

Start by thinking about what new task you will generate. 
Thought: [[pred]]
Answer: [[pred]]
\end{lstlisting}

\begin{lstlisting}[style=textFileStyle, caption=Trajectory Relabeler]
A web-agent is given a precise instruction from a human, which it carries out through a sequence of sub-tasks, where each sub-task (such as clicking on elements / typing on elements / scrolling etc.) changes the HTML state of the webpage.
You are given the initial webpage (as HTML), the final webpage after all sub-tasks are carried out, as well as a summary of changes that each sub-task made to the starting HTML state.

Initial Webpage:
{init_webpage}

Final Webpage:
{final_webpage}

Sub-tasks attempted by the web agent:
{subgoal_str}

Summary of changes made to HTML:
{observation_changes}

Your objective to guess the instruction that was given to the agent. Ensure that your instructions are concrete and such that every sub-task meaningfully contributes to fulfiling the instruction. Start by providing your reasoning. Use the following format for your answer:
Reasoning: your reasoning
Answer: your answer

**Output**
Reasoning: [[pred]]
Answer: [[pred]]
\end{lstlisting}

\begin{lstlisting}[style=textFileStyle, caption=Instruction Following Policy ]
You are a web-agent on an HTML page capable of executing the following kinds of sub-tasks:
{inventory_str}

You are also given some examples of how to perform instructions on the website by converting them into sub-tasks (along with the change each sub-task caused on the website).
{exemplars}

You are given the following instruction: {instruction}. 
To perform this instruction, you've executed the following sub-tasks, and observed the following webpage states (described briefly in language).

**Previous observations and actions**
{prev_observations_and_actions}

After taking these actions, you observe the current web-page HTML:
{webpage_html} 

Webpage Description: [[pred]]

First, think about which inventory item you should pick as your next action.
Thought: [[pred]]

Now, output next action (output *finished* if the instruction has been accomplished) by choosing an item from your inventory
Action: [[pred]]
\end{lstlisting}

\begin{lstlisting}[style=textFileStyle, caption=Demonstration Filter]
You are given an initial web-page from a website (as HTML). To accomplish some task, a web-agent then interacts with the website, leading to a final webpage.
Given the task, the initial webpage and the final webpage, your objective is to judge how well the web-agent carried out this task by giving it a score from 1 to 5.
Only give a score of 5 if the task is perfectly accomplished and the final webpage has no errors.

Task:
{goal_str}

Initial Webpage:
{init_webpage}

Final Webpage:
{final_webpage}

Start by thinking about what the web-agent was trying to accomplish, and describe how well it was done.
Thought: [[pred]]
Answer: [[pred]]
\end{lstlisting}

\subsection{ToolQA} 
\label{sec:toolqa_prompts}
Next, we present all prompts for ToolQA below. The list of methods for various tools in ToolQA is:

\begin{lstlisting}[style=textFileStyle, caption=ToolQA methods ]
(1) Calculate[formula], which calculates the formula and returns the result.
(2) RetrieveAgenda[keyword], which retrieves the agenda related to keyword.
(3) RetrieveScirex[keyword], which retrieves machine learning papers' paragraphs related to keyword.
(4) LoadDB[DBName], which loads the database DBName and returns the database. The DBName can be one of the following: flights/coffee/airbnb/yelp.
(5) FilterDB[condition], which filters the database DBName by the column column_name the relation (e.g., =, >, etc.) and the value value, and returns the filtered database.
(6) GetValue[column_name], which returns the value of the column column_name in the database DBName.
(7) LoadGraph[GraphName], which loads the graph GraphName and returns the graph. The GraphName can be one of the following: PaperNet/AuthorNet.
(8) NeighbourCheck[GraphName, Node], which lists the neighbours of the node Node in the graph GraphName and returns the neighbours. 
(9) NodeCheck[GraphName, Node], which returns the detailed attribute information of Node. 
(10) EdgeCheck[GraphName, Node1, Node2], which returns the detailed attribute information of the edge between Node1 and Node2. 
(11) SQLInterpreter[SQL], which interprets the SQL query SQL and returns the result.
(12) PythonInterpreter[Python], which interprets the Python code Python.
\end{lstlisting}
\vspace{2em}

and the action space for the LM policy is:
\begin{lstlisting}[style=textFileStyle, caption=Action Space ]
(1) Calculate *formula*, which calculates an arithmetic formula (such as 2+3, 2 * 4 etc) and returns the result.
(2) Retrieve passages related to *phrase*, which retrieves information relevant to the supplied phrase. This retriever operates on documents containing information about people's schedules.
(3) Retrieve passages from ML papers related to *keyword*, which retrieves machine learning papers' paragraphs related to keyword.
(4) Load database *DBName*, which loads the database DBName and returns the database. The DBName can be one of the following: flights/coffee/airbnb/yelp.
(5) Filter database according to *condition*. which filters the loaded database (flights/coffee/airbnb/yelp) by a condition and returns the filtered database. A condition is specified as *column_name relation value* where relation can be (=, <, >, <=, >=), and column_name is a column from the loaded DB. To filter according to multiple conditions, the format requires comma separated conditions e.g. "Filter database according to column_name_1=value_1, column_name_2>=value_2, column_name_3<value_3".
(6) Get database value for *column_name*, which returns the value of the column column_name in the database DBName.
(7) Load DBLP, which loads the graphs in dblp. Inside DBLP, there are two graphs: PaperNet/AuthorNet.
(8) List nodes in graph *GraphName*, which lists 10 randomly chosen nodes to help explore the graph.
(9) Check neighbours of node *Node* in graph *GraphName*, which lists the neighbours of the node Node in the graph GraphName and returns the neighbours. GraphName can be PaperNet or AuthorNet.
(10) Get information for node *Node* in graph *GraphName*, which returns the detailed attribute information of Node.
(11) Check edge information between nodes *Node1* and *Node2* in graph *GraphName*, which returns the detailed attribute information of the edge between Node1 and Node2.
(12) Interpret SQLite query: *Query*, which interprets the SQLite query Query and returns the result. There are 4 tables for querying: flights_data/coffee_data/airbnb_data/yelp_data corresponding to the DBs flights/coffee/airbnb/yelp.
(13) Evaluate python code: *code*, which uses the python exec function to execute the python codeblock *code* as is. The result of the code must be stored in a variable called ans, and the code cannot reference any variables not defined inside the codeblock.
(14) Finish with answer: *answer*, which returns the answer and finishes the task.
\end{lstlisting}
\vspace{2em}

This is then directly used for various prompts as \texttt{\{inventory\_str\}}. Note that the action strings (from this inventory) are converted into actual methods via string post-processing.

\begin{lstlisting}[style=textFileStyle, caption=Exploration Policy]
You are an agent with access to tools, that you may use to respond to various questions. You have the following tools:
{inventory_str}

Your objective is to discover diverse and interesting questions (that a human might give to an agent with these tools) by chaining together calls to different tools. You've executed the following tool calls, and observed the following outputs from these tools (described briefly in language).

**Previous observations and actions**
{prev_observations_and_actions}

**Current Observation**
{curr_observation}

Start by thinking about what action you should take next.
Thought: [[pred]]

Now, act by taking an action based in the inventory (or output Finish if you are done).
Action: [[pred]]
\end{lstlisting}

\begin{lstlisting}[style=textFileStyle, caption=Instruction Generator]
**Objective**
You are an agent with access to tools, that you may use to respond to various queries. You have the following tools:
{inventory_str}

To respond to queries, you need to call tools in a specific sequence to obtain the answer.

Your objective is to propose a query that can be performed by chaining together these tools. Ensure that your queries are concrete. 

Start by thinking about what new query you will generate. 
Thought: [[pred]]
Answer: [[pred]]
\end{lstlisting}

\begin{lstlisting}[style=textFileStyle, caption=Trajectory Relabeler]
A user asks an AI agent a question, which it answers by accessing tools like databases, calculators, retrievers and python interpreters. The AI agent answers this question by carrying out a sequence of sub-tasks, where each sub-task (such as loading or querying a dblp graph / calling a python interpreter etc.) leads to an output from the tool. 

You are given the entire sequence of tool outputs, where the final tool output is the answer that the agent gives. You are also given the sequence of sub-tasks attempted by the agent.

Sub-tasks attempted by the agent:
{subgoal_str}

Sequence of tool outputs:
{observation_changes}

Your objective to guess the query that was given to the agent. Ensure that your answer is concrete and such that every sub-task meaningfully contributes to answering the query. Start by providing your reasoning. Use the following format for your answer:
Reasoning: your reasoning
Answer: your answer

**Output**
Reasoning: [[pred]]
Answer: [[pred]]
\end{lstlisting}

\begin{lstlisting}[style=textFileStyle, caption=Instruction Following Policy ]
You are an agent with access to tools, that you may use to respond to various queries. You have the following tools:
{inventory_str}

To respond to queries, you need to call tools in a specific sequence to obtain the answer. Here are some demonstrations of how to respond to queries by invoking tools:
{exemplars}

You are given the following query: {super_goal} 

To perform this instruction, you've executed the following actions, and observed the following outputs from your tools: 

**Previous observations and actions**
{prev_observations_and_actions}

**Current Observation**
{curr_observation}

First, think about which tool you should pick as your next action
Thought: [[pred]]

Now, output next action (output *finished* if the instruction has been accomplished) by calling the chosen tool with appropriate arguments. End your output with a newline
Action: [[pred]]
\end{lstlisting}

\begin{lstlisting}[style=textFileStyle, caption=Demonstration Filter]
A user asks an AI agent a question, which it answers by accessing tools like databases, calculators, retrievers and python interpreters. The AI agent answers this question by carrying out through a sequence of sub-tasks, where each sub-task (such as loading or querying a dblp graph / calling a python interpreter etc.) leads to an output from the tool. You are given the entire sequence of tool outputs, where the final tool output is the answer that the agent gives. You are also given the sequence of sub-tasks attempted by the agent.

Your objective is to judge how well the AI agent carried out this task by giving it a score from 1 to 5.
Only give a score of 5 if the task is perfectly accomplished and the final answer has no errors.

User question:
{goal_str}

Sequence of Tool outputs:
{state_changelog}

Start by thinking about what the AI agent was trying to accomplish, and describe how well it was done.
Thought: [[pred]]
Answer: [[pred]]
\end{lstlisting}

\section{Converting LM Action space into API calls}
\label{sec:lm_actions_to_api}

\paragraph{MiniWoB++.} We use the following prompt to convert the action string into an API call: 
\begin{lstlisting}[style=textFileStyle, caption=LM to convert action strings into an API call]
Webpage HTML: {html}
Use references into the webpage to specify actions to perform a given task.

You can take 4 kinds of actions on a chosen element specified via its ref id.

Action: type(text) types 'text' into chosen ref, useful for typing into various textboxes.
Action: click() clicks on chosen element, useful when clicking buttons,checkboxes or textboxes. Sections can be clicked for expansion.
Action: move-mouse() moves mouse to a chosen element, useful when the element text has '>' symbol for expansion.
Action: clear() clears all text on chosen ref-id, useful when you want to delete text on textboxes.


To choose actions, strictly use the format below:
Chosen action: chosen from click/move-mouse/type/clear
Chosen element: Specify chosen ref id as an integer
Chosen text: text to type (n/a if chosen action is not type)

Task: {action_string}
Chosen action: [[pred]]
Chosen element: [[pred]]
Chosen text: [[pred]]
\end{lstlisting}

\vspace{2em}
The LM predictions are combined into an API call e.g. \texttt{ref[[element]].type([[text]]])}. We use a simple python function to convert the API call into a Selenium web-driver method (\textit{type\_text}, \textit{clear} and \textit{move\_mouse} are Selenium web-driver methods):

\section{Comparing BAGEL Agents with other Few-shot agents}
\label{sec:comparisons}

In Table~\ref{tab:agent_comparison}, we compare $\ours{}$ agents with recently proposed zero-shot and few-shot agents for MiniWoB++. Specifically, we compare with the ``Flat Zero-Shot'' and ``SteP Zero-Shot'' agents from \citet{sodhi2023heap} and RCI \citep{kim2023language}. We provide these just as reference, noting that results are not entirely comparable due to different underlying language models.

\begin{table*}[h!]
  \centering
  \small
\renewcommand{\arraystretch}{1.2}
  \begin{tabular}{@{}lcccc@{}}
    \toprule
    Task & BAGEL (PaLM-2) & Flat Zero-Shot (GPT 3.5) & SteP Zero-Shot (GPT 3.5) & RCI ( GPT-4) \\ 
    \midrule

    book-flight           &  \textbf{0.15}   &  0.0 &  0.0   & -   \\ 
    choose-date           & \textbf{0.4}    &  0.2  &  0.2 &  -    \\ 
    social-media          & 0.7    &   - &   - & \textbf{1.0}    \\ 
    email-inbox           &  \textbf{1.0}   &  0.4  & 0.0    &  0.98     \\ 
    click-checkboxes-soft & \textbf{0.9}    &  0.0  & 0.04 &   0.72    \\ 
    click-tab-2-hard      &  \textbf{1.0}   &  0.68  & 0.76  & 0.76       \\ 
    social-media-some     & 0.8    &  -  &  -  &  \textbf{0.9}     \\ 
    tic-tac-toe           &  0.4   &  -   & - &   \textbf{0.56}    \\ 
    use-autocomplete      &  0.45   &  -   &  -  & \textbf{0.58}      \\ 
    search-engine         & 0.25    &  0.38  & 0.26  &  \textbf{1.00}     \\ 
    \bottomrule
  \end{tabular}
  \caption{Comparising BAGEL Agents with other agents for MiniWoB++. We provide these results as reference and note that the underlying language models are different.}
  \label{tab:agent_comparison}
\end{table*}

\end{document}